\documentclass{ieeeaccess}
\usepackage{cite}
\usepackage{amsmath,amssymb,amsfonts}
\usepackage{algorithmic}
\usepackage{graphicx}
\usepackage{xcolor}
\usepackage{caption}
\usepackage{subcaption}
\usepackage{multicol}
\usepackage{textcomp}
\usepackage{blindtext}
\usepackage{hyperref}

\newtheorem{remark}{Remark}

\def\BibTeX{{\rm B\kern-.05em{\sc i\kern-.025em b}\kern-.08em
    T\kern-.1667em\lower.7ex\hbox{E}\kern-.125emX}}
\begin{document}
\history{Date of publication xxxx 00, 0000, date of current version xxxx 00, 0000.}
\doi{10.1109/ACCESS.2017.DOI}

\title{The STDyn-SLAM: a stereo vision and semantic segmentation approach for SLAM in dynamic outdoor environments}
\author{\uppercase{Daniela Esparza},
\uppercase{Gerardo Flores}}
\address{Centro de Investigaciones en Óptica (CIO),
Lomas del Bosque 115, Lomas del Campestre, 37150, León, Guanajuato. (e-mail: daniesparza@cio.mx, gflores@cio.mx)}

\tfootnote{This work was supported in part by the project FORDECYT-CONACYT 292399.}

\markboth
{D. Esparza \headeretal: A Stereo Vision and Semantic Segmentation Approach for SLAM in Dynamic Outdoor Environments}
{D. Esparza \headeretal: A Stereo Vision and Semantic Segmentation Approach for SLAM in Dynamic Outdoor Environments}

\corresp{Corresponding author: Gerardo Flores.}


\begin{abstract}
The Visual Simultaneous Localization and Mapping (SLAM) is a system based on the scene's features to estimate a map and the system pose. Commonly, SLAM algorithms are focused on a static environment; however, some dynamic objects are present in the vast majority of real-world applications. This work presents a feature-based SLAM system focused on dynamic environments using convolutional neural networks, optical flow, and depth maps to detect objects in the scene. The proposed system employs a stereo camera as the primary sensor to capture the scene. The neural network is responsible for object detection and segmentation to avoid erroneous maps and wrong system locations. Moreover, the proposed system's processing time is fast and can run in real-time, running in outdoor and indoor environments. The proposed approach has been compared with state-of-the-art; besides, we present several experimental results outdoors that corroborate the approach's effectiveness. Our code is available online.
\end{abstract}


\titlepgskip=-15pt

\maketitle

\section*{Supplementary material}
The implementation of our system is released on GitHub and is available under the following link: \newline
\url{https://github.com/DanielaEsparza/STDyn-SLAM} 

Besides, this letter has supplementary video material available at \url{https://youtu.be/3tnkwvRnUss}, provided by the authors. 

\section{Introduction} \label{sec:introduction}
Simultaneous Localization and Mapping (SLAM) systems are strategic for the development of the following navigation techniques. This is mainly due to its fundamental utility in solving the problem of autonomous exploration tasks in unknown environments such as mines, highways, farmlands, underwater/aerial environments, and in broad terms, indoor and outdoor scenes. The problem of SLAM for indoor environments has been investigated for years, where usually RGB-D cameras or Lidars are the primary sensors to capture scenes \cite{795798}, \cite{844732}, \cite{fastslam}. Indoors, dynamic objects are usually more controllable, unlike outdoors, where dynamic objects are inherent to the scene.

\begin{figure}[t]
    \centering
    \subcaptionbox{Left}{\includegraphics[width=0.33\columnwidth]{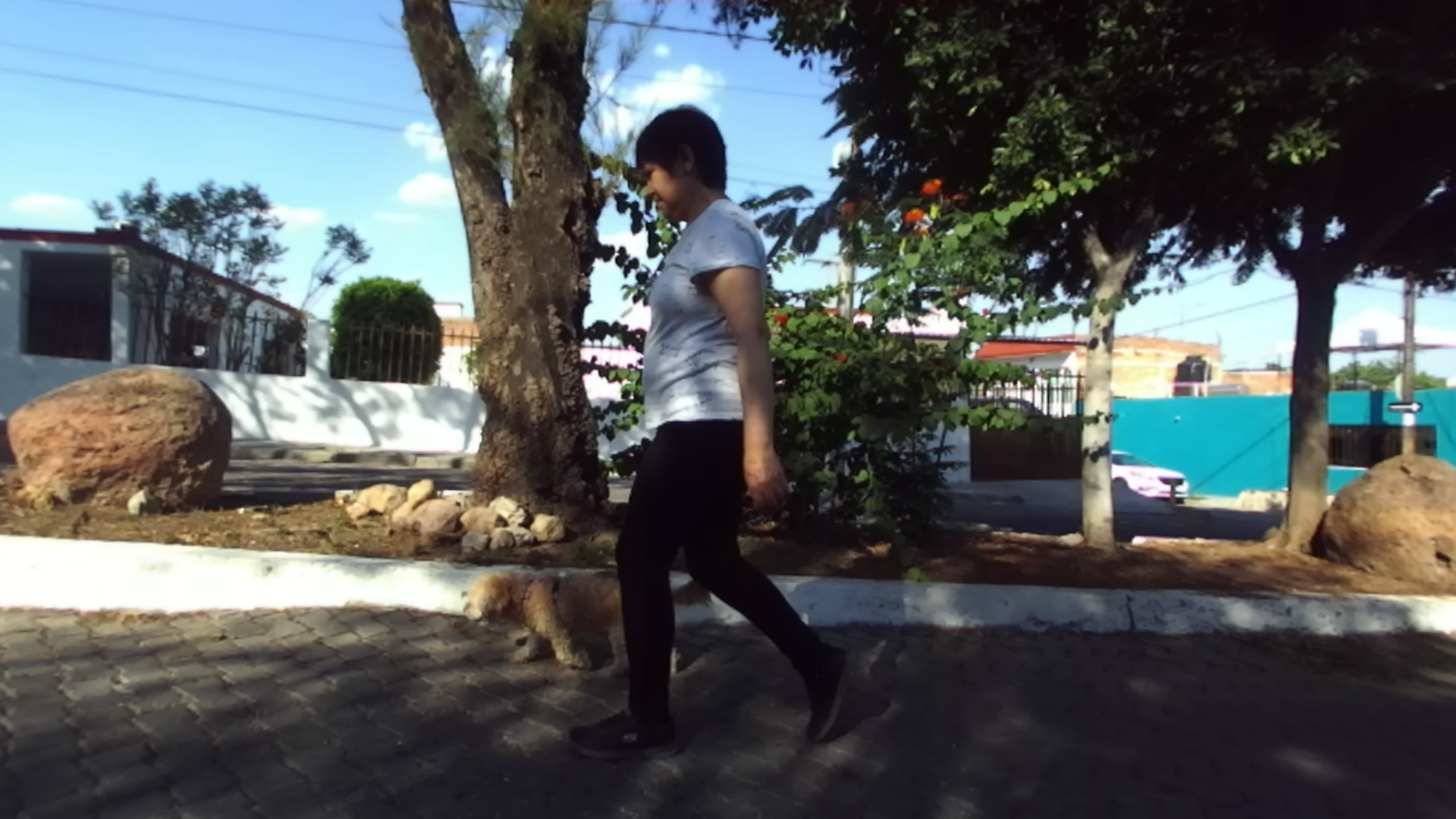}}%
    \hfill 
    \subcaptionbox{Right}{\includegraphics[width=0.33\columnwidth]{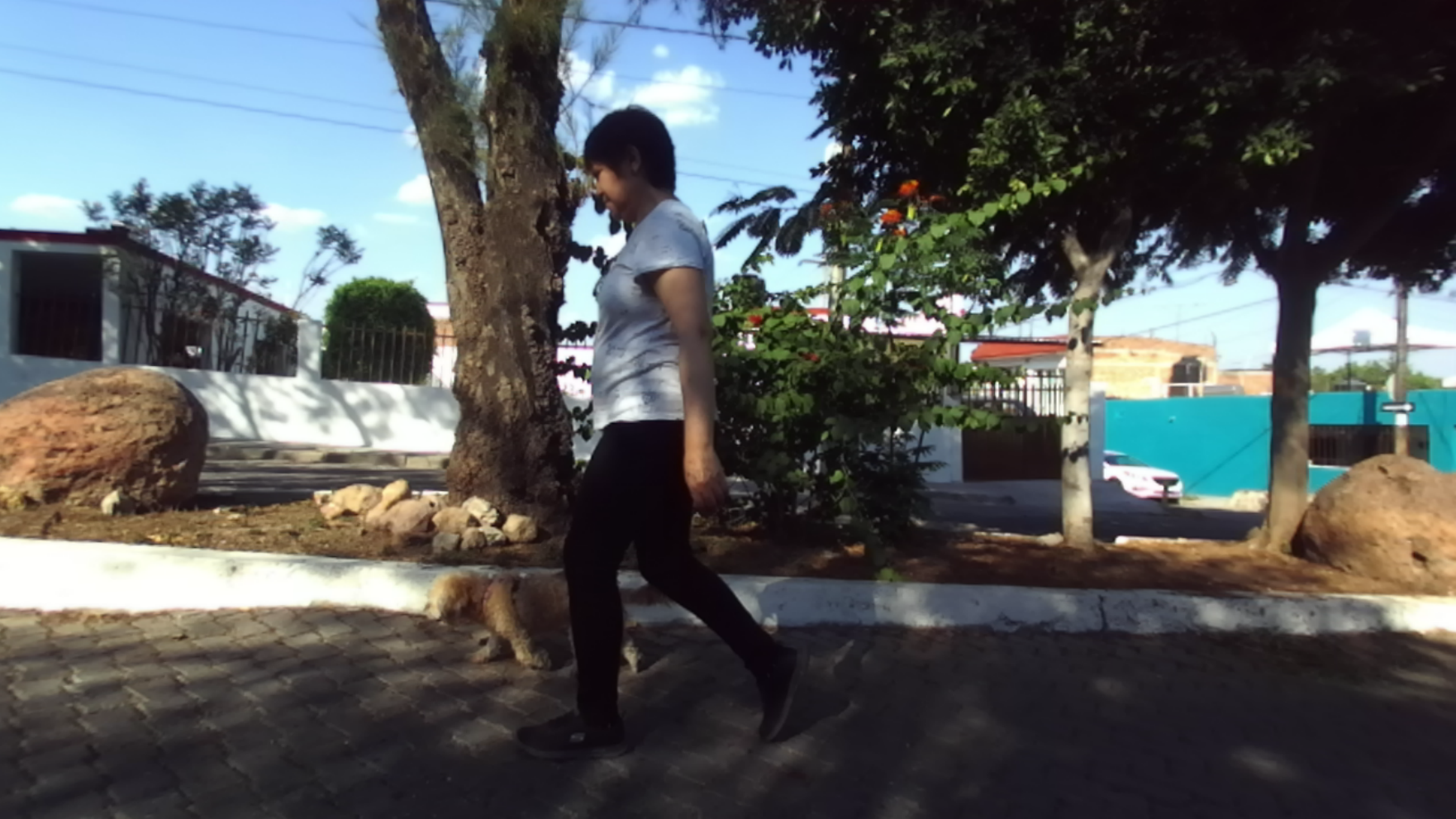}}%
    \hfill 
    \subcaptionbox{Depth}{\includegraphics[width=0.33\columnwidth]{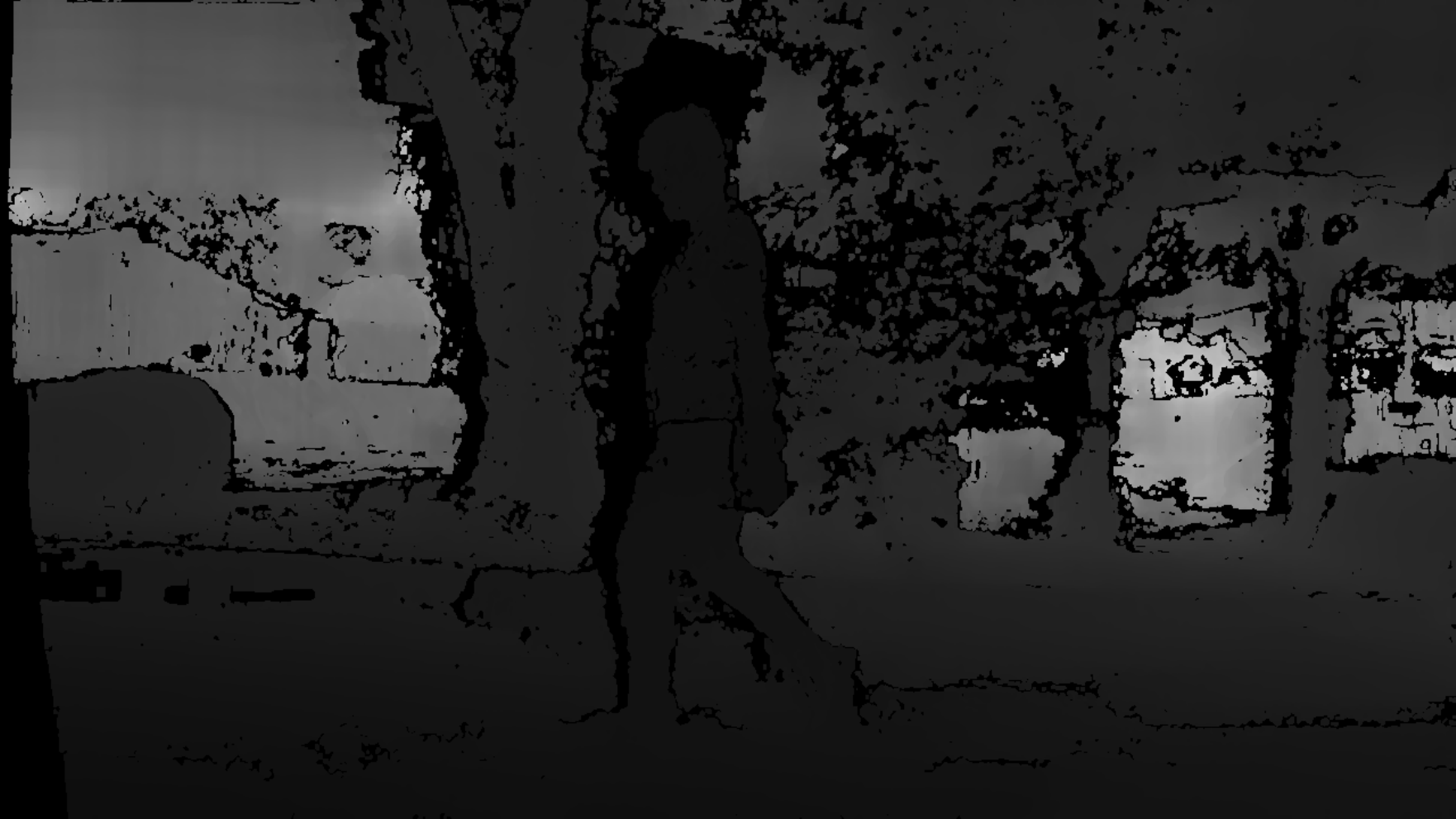}}%
    \\
    \subcaptionbox{3D reconstruction \label{abstract_d} }{\includegraphics[width=\columnwidth]{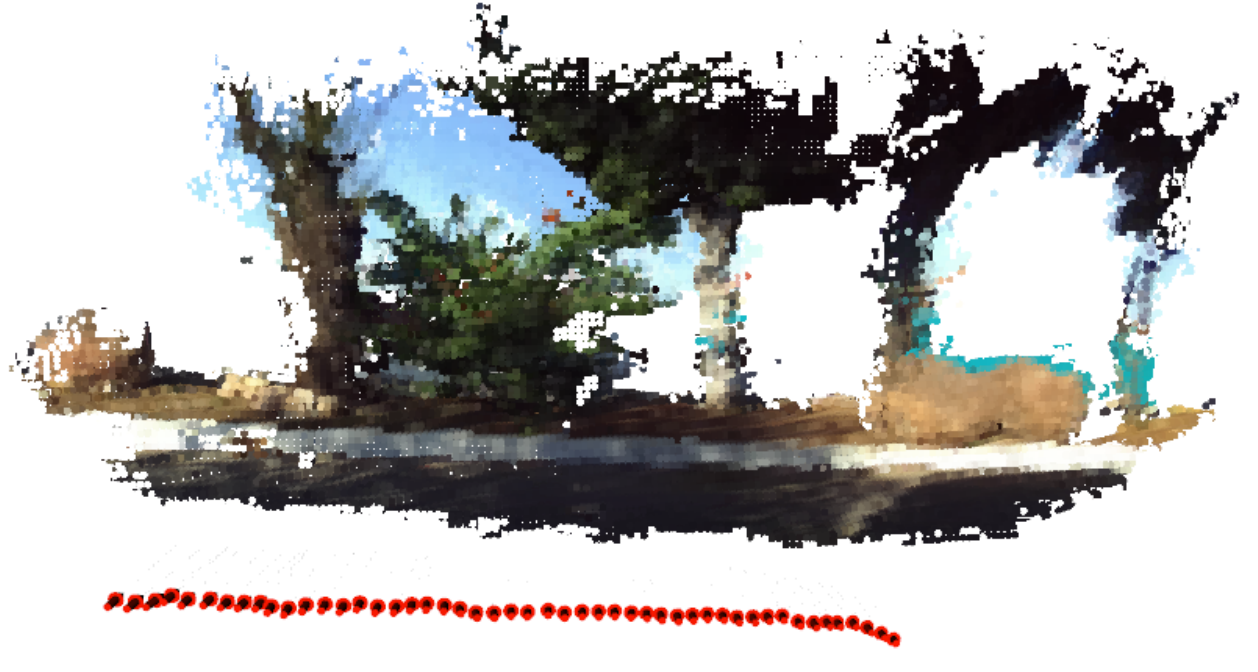}}
    \caption{The STDyn-SLAM results in scenes with moving objects. First raw: Input images with two dynamic objects. Second raw: 3D reconstruction performed by the STDyn-SLAM discarding moving objects.}
    \label{fig:abstract}
\end{figure}

On the other hand, the vast majority of SLAM systems are focused on the static environment assumption, such as HECTOR-SLAM \cite{HECTOR_SLAM}, Kintinuous \cite{Whelan-2012-7552}, MonoSLAM \cite{4160954}, PTAM \cite{Yoshinari-Kameda2012}, SVO \cite{svo}, LSD-SLAM \cite{LSD-SLAM} among others. Such an assumption is strong since it restricts the SLAM system to work only in static environments. However, in dynamic environments, the moving objects can generate an erroneous map and wrong poses because dynamic features cause a bad pose estimation and erroneous data. For this reason, new approaches have arisen for solving the dynamic environment problem. There are few systems in the last two years focused on SLAM for outdoor environments, such as NeuroSLAM \cite{NeuroSLAM}, hierarchical Outdoor SLAM \cite{5169855}, and Large-Scale Outdoor SLAM \cite{large-scale}.


\begin{table*}[t]
    \centering
    \caption{The state of the art of the SLAM problem considering dynamic environments.}
    \begin{tabular}{c | c | c | c | c | c }
         \hline
        & & & &\\
        System & Sensor & Environment & Dynamic Objects & Real Time & Method\\[5pt]
        \hline
        \hline
        & & & &\\
        
        \cite{sof}                  & RGB-D   & Indoor &   PASCAL VOC    & $-$ & Semantic segmentation, \\ 
        &         &        &                 &     & and optical flow  \\[5pt]
        
        \cite{dsod}                 &   Mono  & Indoor &       COCO      & $-$ & Semantic segmentation network,   \\
                                    &         & Outdoor& &                      & depth prediction network  \\  
                                    &         &        &                 &     & and geometry properties  \\[5pt] 
        
        \cite{cubeslam}             &   Mono  & Indoor & YOLO and  & GPU &  3-D Box Proposal Generation,  \\
                                    &         & Outdoor&          MS-CNN       &     & standard 3-D map point  \\
                                    &         &        &                 &     & reprojection error, constant motion   \\
                                    &         &        &                 &     & model with uniform velocity  \\[5pt] 
        
        \cite{rgbd_dyn}             &  RGB-D  & Indoor &       COCO      & $-$ & Mask R-CNN, edge refinement,     \\ 
                                     &         &        &                 &     &  and optical flow \\[5pt] 
        
        \cite{Henein2020DynamicST}  &  RGB-D/Stereo  & Indoor &    COCO    & $-$ &  Factor graph and instance-level object \\  
                                    &   and proprioceptive  & Outdoor &     &  & segmentation algorithm \\[5pt]
                                    
        Ours  &  RGB-D/Stereo  & Indoor &    PASCAL VOC     & GPU & Semantic segmentation, optical flow   \\ 
                                    &    & Outdoor &     &  &  and epipolar geometry \\[5pt]

        \hline
    \end{tabular}
    \label{tabla:state-of-art}
    
\end{table*}

In this work, we propose a method called \textit{STDyn-SLAM} for solving SLAM's problem in dynamic outdoor environments using stereo vision \cite{8798272}. Fig. \ref{fig:abstract} depicts a sketch of our proposal in real experiments. The first row shows the input images, where a potentially dynamic object is present on the scene and is detected by a semantic segmentation neural network. Fig. \ref{abstract_d} depicts the 3D reconstruction excluding dynamic objects. To evaluate our system, we carried out experiments in different outdoor scenes, and we qualitatively compare the 3D reconstructions taking into account the excluding of dynamic objects. We conducted experiments using sequences from KITTI Dataset, and they are compared with state-of-the-art systems.
Furthermore, our approach is implemented in ROS, which facilitates implementations in various applications. Also, the STDyn-SLAM approximately works to 10 frames per second using ROS or offline from stored images. Further, we publish our code been available on GitHub.\footnote{\underline{https://github.com/DanielaEsparza/STDyn-SLAM}} Also, a video is available on Youtube.

The rest of the paper is structured as follows. Section \ref{Related_work} presents the related work of SLAM in dynamic environments. In Section \ref{Methods}, we present the main results and the algorithm STDyn-SLAM algorithm. Section \ref{sec:experiments} presents the real-time experiments of STDyn-SLAM in outdoor environments with moving objects; we compare our approach with state-of-art methods using the KITTI dataset. Finally, the conclusions and the future work are given in Section \ref{conclusion}. 
\begin{figure*}[t]
    \centering
    \includegraphics[width=500pt]{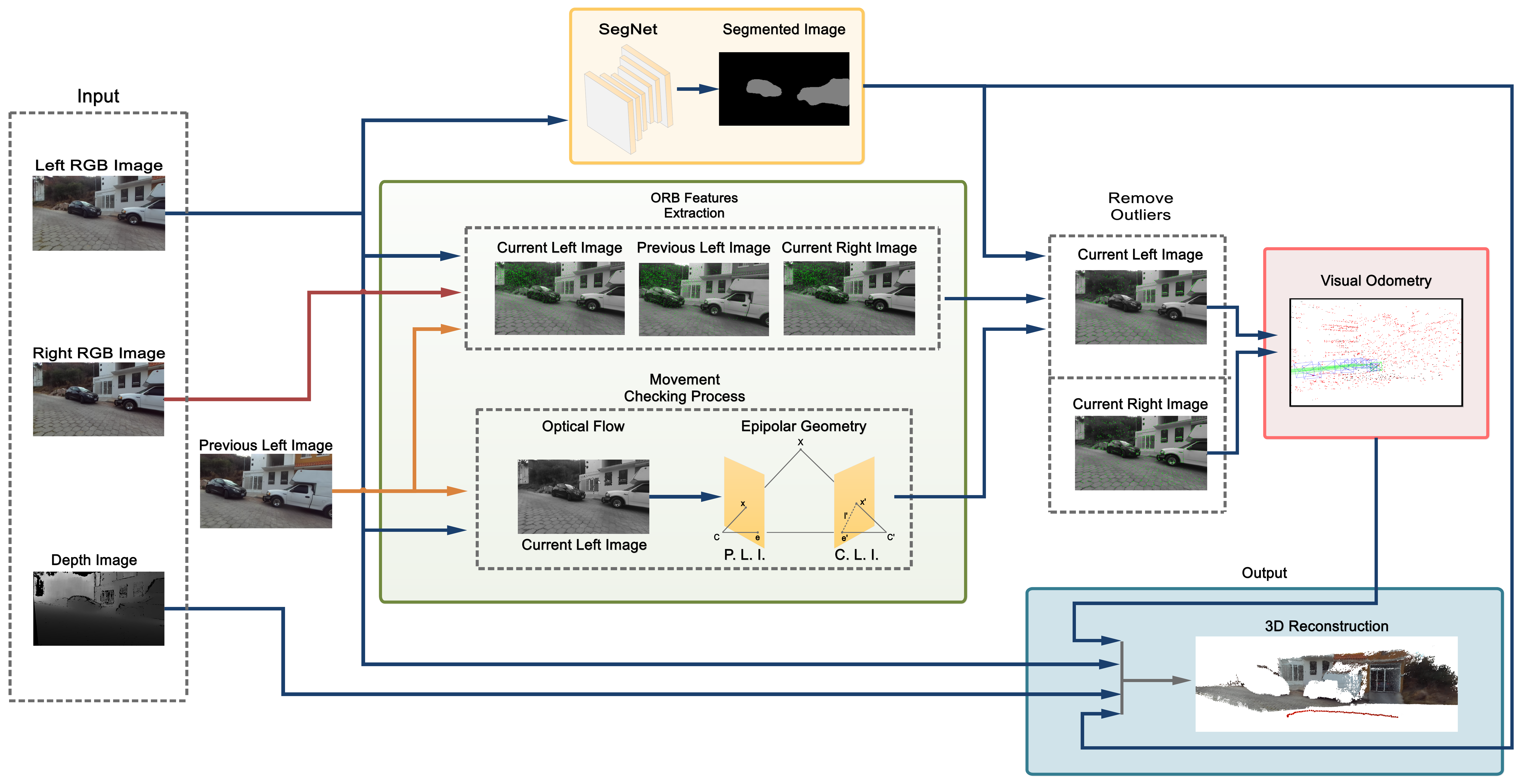}
    \caption{A block diagram showing the algorithm steps of the STDyn-SLAM.}
    \label{fig:stereo_process}
\end{figure*}

\section{Related work}\label{Related_work}
\subsection{Classic approaches}
The classical methods do not consider artificial intelligence. Some of these approaches are based on optical-flow, epipolar geometry, or a combination of the two. In \cite{DRESLAM}, Yang \textit{et al.} propose a SLAM system using an RGB-D camera and two encoders for estimating the pose and building an OctoMap. The dynamic pixels are removed using an object detector and a K-means to segment the point cloud. On the other hand, in \cite{Gimenez2019}, Gimenez \textit{et al.} present a CP-SLAM based on continuous probabilistic mapping and a Markov random field; they use the iterated conditional modes. Wang et al. \cite{Wang2019} propose a SLAM system for indoor environments based on an RGB-D camera. They use the number of features on the static scene and assume that the parallax between consecutive images is a constraint of movement. In \cite{flow_dynamic}, Cheng, Sun, and Meng implement an optical-flow and the five-point algorithm approach to obtain dynamic features. In \cite{optical_ransac}, Ma and Jia proposed a visual SLAM for dynamic environments detecting the dynamic objects in the scene using optical flow. Furthermore, they use the RANSAC algorithm to improve the computation of the homography matrix.

\subsection{Artificial-intelligence-based approaches}
Thanks to the growing use of deep learning, in the last three years some SLAM systems use artificial-intelligence-based approaches. Table \ref{tabla:state-of-art} resumes the state-of-art in this regard. Some of the works such as Dosovitskiy \textit{et al.} \cite{flow_dosovitskiy}, Ilg \textit{et al.} \cite{flownet_2_0} and Mayer \textit{et al.} \cite{scene_flow_mayer} use optical flow together with supervised learning for detecting and segmenting the moving objects. 

In \cite{mid-fusion}, Xu \textit{et al.} proposed an instance segmentation of the objects in the scene based on the COCO dataset \cite{Coco}. The geometric and motion properties are detected and used to improve the mask boundaries. Also, they tracked the visible objects and moving objects and estimate the system's pose. Several works are based on RGB-D cameras such as \cite{sof} , \cite{rgbd_dyn}, and \cite{Henein2020DynamicST}. Linyan Cui and Chaowei Ma \cite{sof} proposed the SOF-SLAM, an RGB-D system based on ORB-SLAM2, which combines a neural network for semantic segmentation, and optical flow for removing dynamic features. Lili Zhao et al. \cite{rgbd_dyn} proposed an RGB-D framework to dynamic scenes, where they combined the Mask R-CNN, edge refinement, and optical flow to detect the probably dynamic objects. Mina Henein et al. \cite{Henein2020DynamicST} proposed a system based on an RGBD camera and proprioceptive sensors, where they tackled the SLAM problem with a model of factor graph and an instance-level object segmentation algorithm to the classification of objects and the tracking of features. The proprioceptive sensors are used to estimate the camera pose. Also, there are works using a monocular camera, for instance, the DSOD-SLAM presented in \cite{dsod}. Ping Ma et al. employs a semantic segmentation network, a depth prediction network, and geometry properties to improve the results in dynamic environments.  Our work is built on the well-known ORB-SLAM2 \cite{ORB-SLAM2} taking some ideas from DS-SLAM system \cite{DS-SLAM}. In the DS-SLAM, the authors used stored images from an RGB-D camera for solving the SLAM problem in indoor dynamic environments. Nevertheless, the depth map typically obtained from an RGB-D camera is hard to get for external environments.


\section{Methods}\label{Methods}
In this section, we present and describe the framework of the STDyn-SLAM with all the parts that compose it. A block diagram describing the framework's pipeline is depicted in Fig. \ref{fig:stereo_process}, where the inputs at the instant time $t$ are the stereo pair, depth image, and the left image captured at $t-1$ (aka previous left image). The process starts with the extraction of ORB features in the stereo pair and the past left image. Then, it follows the optical flow and epipolar geometry image processing. In parallel, the neural network segments potentially moving objects in the current left image. To remove outliers (features inside dynamic objects) and estimate the visual odometry, it is necessary to computation the semantic information and the movement checking process. Finally, the 3D reconstruction is computed from the segmented image, visual odometry, the current left frame, and the depth image. These processes are explained in detail in the following subsections.

\subsection{Stereo Process}
Motivated by the vast applications of robotics outdoors, where dynamic objects are presented, we proposed that our STDyn-SLAM system be focused on stereo vision. A considerable advantage of this is that the depth estimation from a stereo camera is directly given as a distance measure. The process described in this part is depicted in Fig.  \ref{fig:stereo_process}, where three main tasks are developed: feature extraction, optical flow, and epipolar geometry. Let begin with the former. 

The first step of the stereo process is acquiring the left, right, and depth frames from a stereo camera. Then, a local feature detector is applied in the stereo pair and the previous left image. As a feature detector, we use the Oriented fast and Rotated Brief (ORB) feature detector, which throws the well-known ORB features \cite{6126544}. Once the ORB features are found, optical flow and a process using epipolar geometry is conducted.

To avoid dynamic objects not classified by the neural network (explained in the following subsection), the STDyn-SLAM computes optical flow using the previous and current left frames. This step employs a Harris detector to compute the optical flow with its features different from the ORB ones. The Harris points pair is discarded if at least one of the points is on the edge corner or close to it.

From the fundamental matrix, ORB features, and optical flow, we compute the epipolar lines. Thus, we can map the matched features from the current left frame into the previous left frame. The distance from the corresponding epipolar line to the mapped feature into the past left image determines an inlier or outlier. 

\subsection{Artificial neural network's architecture}\label{network_architecture} 
The approach we use is that of eliminating the ORB features on dynamic objects. For that, we need to discern the real dynamic objects among all the objects in the scene. It is here where the NN depicted in Fig.  \ref{fig:stereo_process} is introduced. In the NN block of that figure, a semantic segmentation neural network is shown, with the left image as input and a segmented image with the object of interest as output. This NN is a pixel-wise classification and segmentation framework. The STDyn-SLAM implements a particular NN of this kind called SegNet \cite{segnet}, which is an encoder-decoder network based on the VGG-16 network \cite{vgg16}. The encoder of this NN architecture counts with thirteen convolutional layers with batch normalization, a ReLU non-linearity divided into five encoders, and five non-overlapping max-pooling and sub-sampling layers located at the end of each encoder. Since each encoder is connected to a corresponding decoder, the decoder architecture has the same number of layers as encoder architecture, and every decoder has an upsampling layer at first. The last layer is a softmax classifier. SegNet classifies the pixel-wise using a model based on the PASCAL VOC dataset \cite{pascal}, which consists of twenty classes. The pixel-wise can be classified into one of the following classes: airplane, bicycle, bird, boat, bottle, bus, car, cat, chair, cow, dining table, dog, horse, motorbike, person, potted plant, sheep, sofa, train and TV/monitor.

Notwithstanding those above, not all feature points in the left frame are matched in the right frame. For that reason and for saving computing resources, the SegNet classifies the objects of interest only on the left input image.

\subsubsection{Outliers removal} 
\begin{figure}
    \centering 
    \includegraphics[width=\columnwidth]{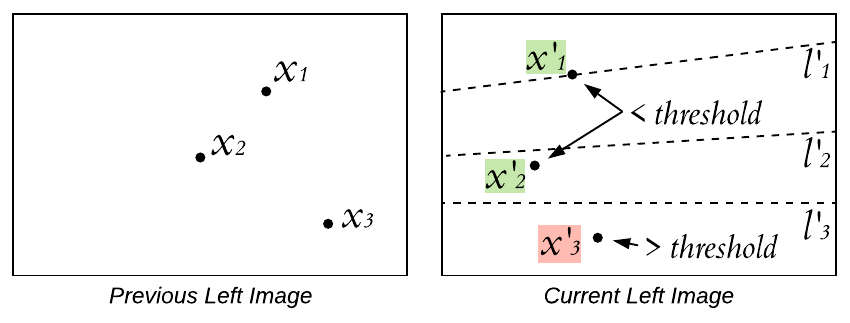}
    \caption{The cases of inliers and outliers. Green: the $x_1'$ and $x_2'$ are inliers; the distance from the point to their corresponding epipolar line $l'$ is less than a threshold. Red: $x_3'$ is an outlier since the distance is greater than the threshold.}
    \label{fig:distance}
\end{figure}

Once the all the previous steps have accomplished, a threshold is selected to determine the features as inlier or outlier. Fig. \ref{fig:distance} depicts the three cases of a mapped feature. Let $x_1$, $x_2$, and $x_3$ denote the ORB features from the previous left image; $x_1'$, $x_2'$, and $x_3'$ denote the corresponding features from the current left image; and $l_1'$, $l_2'$, and $l_3'$ are the epipolar lines. The first and second cases correspond to inliers, $x_1'$ is over $l_1'$, and the distance from $x_2'$ to $l_2'$ is less than the threshold. The third case is an outlier because the distance from $x_3'$ to $l_3'$ is greater than the threshold. 

Remember that the SegNet, described before, semantically segments the left image in object classes. The semantic segmentation enhances the rejection of ORB features on the possible dynamic objects. The ORB features inside segmented objects, and thus possible moving objects, are rejected. The remained points are matched with the ORB features from the right image.

\subsection{Visual odometry}
Because the system is based on ORB-SLAM2, the SLAM visually computes the odometry. Therefore, the next step needs the ORB features to estimate the depth for each feature pair. The features are classified in mono and stereo and will be necessary to track the camera's pose. This step is merely a process from ORB-SLAM2.

\subsection{3D reconstruction}
Finally, the STDyn-SLAM builds a 3D reconstruction from left, segmented, and depth images using visual odometry. First, the 3D reconstruction process checks each pixel of the segmented image to reject the point corresponding to the classes of the objects selected as dynamic in section \ref{network_architecture}. Then, if the pixel is not considered as a dynamic object, the equivalent pixel from the depth image is added to the point cloud, and the assigned color of the point is obtained from the left frame. This section builds a local point-cloud only in the current pose of the system, and then the octomap \cite{Hornung12octomap:an} joins and updates the local point clouds in a full point cloud.

\begin{remark}
It is essential to mention that we merely applied the semantic segmentation, optical flow, and geometry constraints to the left image to avoid increasing the time executing. Moreover, the right frame segmentation is unnecessary because feature selection rejects the ORB features inside dynamic objects from the left image, so the corresponding points from the right frame will not be matched.
\end{remark}

\begin{figure}[t]
    \centering
    \includegraphics[width=0.85\columnwidth]{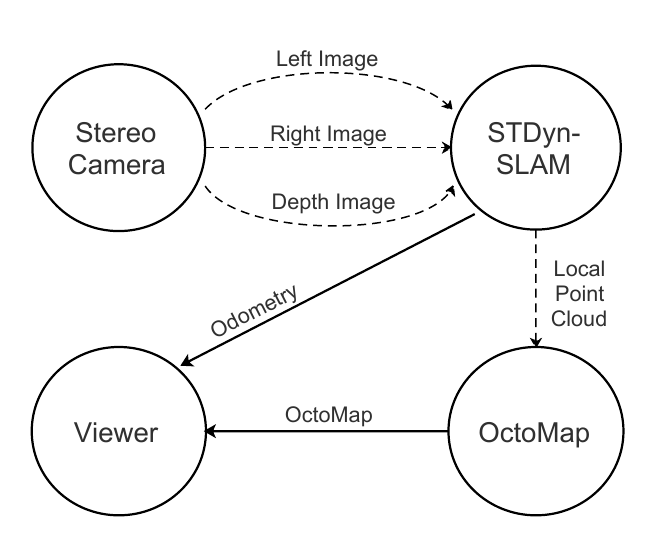}
    \caption{Diagram of the ROS nodes of the STDyn-SLAM required to generate the trajectory and 3D reconstruction. The circles represent each process's ROS node, and the arrows are the ROS topics published by the ROS nodes. The continued arrows depict the final ROS topics.}
    \label{fig:ros_process}
\end{figure}


\section{Experiments} \label{sec:experiments}
In this section, we test our algorithm STDyn-SLAM in real-time scenes under the KITTI datasets. Our system's experiments were compared to other state-of-art systems to evaluate the 3D reconstruction and the odometry. The results of the 3D map were qualitatively measured because of the nature of the experiment. We employ the Absolute Pose Error (APE) metric for the odometry.

\subsection{Hardware and software setup}
We tested our system on an Intel Core i7-7820HK laptop computer with 32 Gb RAM and a GPU GeForceGTX 1070. Moreover, we used as input a ZED camera, which is a stereo camera developed by Sterolabs. We selected an HD720 resolution. The ZED camera resolutions are WVGA $(672\times376)$, HD720 $(1280\times720)$, HD1080 $(1920\times1080)$ and 2.2K $(2208\times1242)$.

STDyn-SLAM is developed naturally on ROS. Our system's main inputs are the left and right images, but the depth map is necessary to build the point cloud. However, if this is not available, it is possible to execute the STDyn-SLAM only with the stereo images and then obtain the trajectory. On the other hand, the \textit{STDyn} node in ROS generates two main topics; the \textit{Odom} and the \textit{\small{ORB$\_$SLAM2$\_$PointMap$\_$SegNetM$/$Point$\_$Clouds}} topics. The point cloud topic is the input of the \textit{octomap$\_$server} node; this node publishes the joined point cloud of the scene. 

Fig. \ref{fig:ros_process} depicts the required ROS nodes by the STDyn-SLAM to generate the trajectory and the 3D reconstruction. The camera node publishes the stereo images and computes the depth map from them. Then, the STDyn-SLAM calculates the odometry and the local point cloud. The OctoMap combines and updates the current local point cloud with the previous global map to visualize the global point cloud. It is worth mentioning that the user can choose the maximum depth of the local point cloud. All the ROS topics can be shown through the viewer.

\subsection{Real-time experiments}
We present real-time experiments under three different scenarios explained next.

\begin{figure}[t]
    \centering
    \subcaptionbox{First image of the sequence.}{\includegraphics[width=0.495\columnwidth]{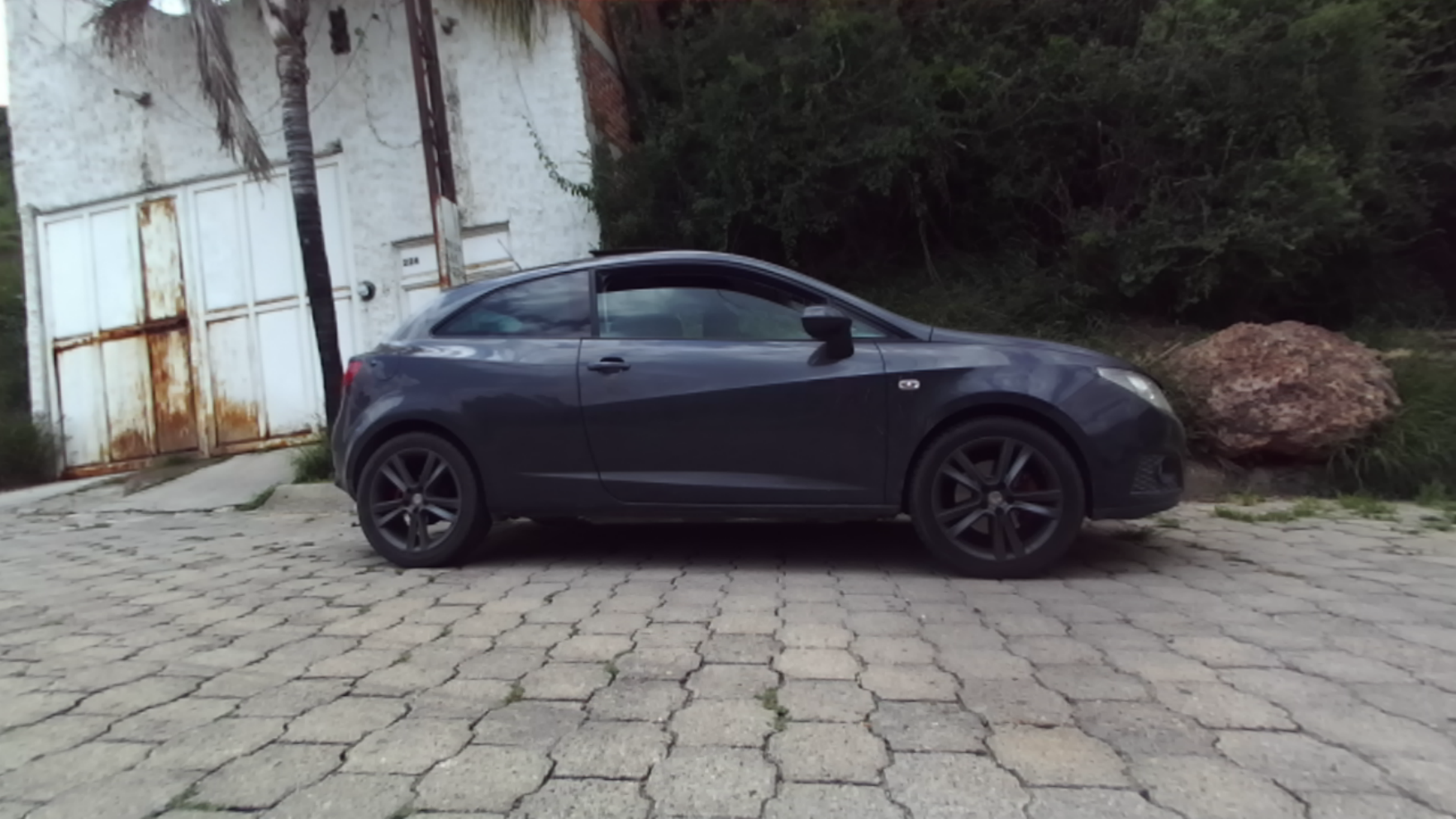}}%
    \hfill 
    \subcaptionbox{Seventh image of the sequence.}{\includegraphics[width=0.495\columnwidth]{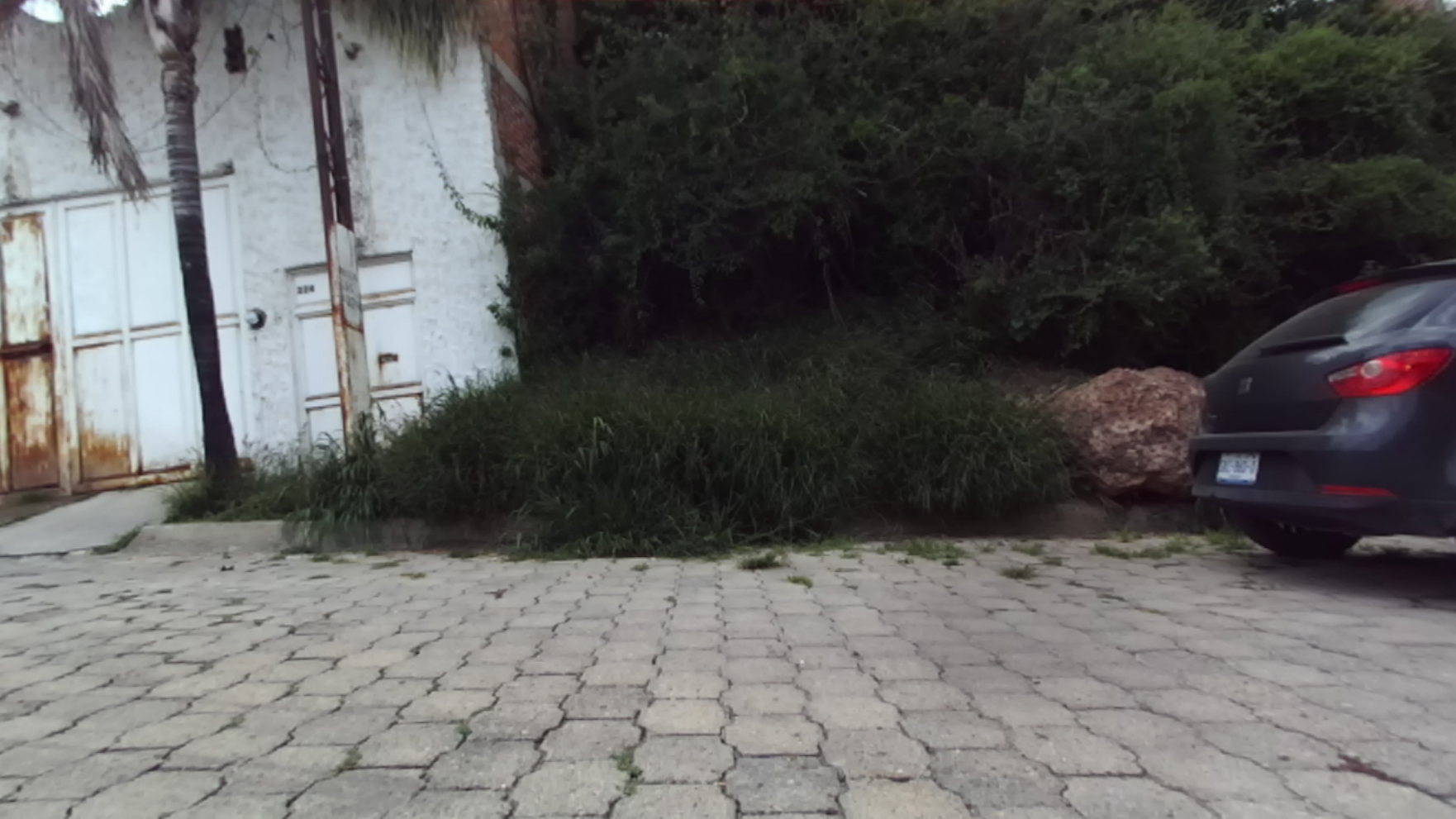}}%
    \\
    \subcaptionbox{3D reconstruction. \label{ibiza_3d} }{\includegraphics[width=\columnwidth]{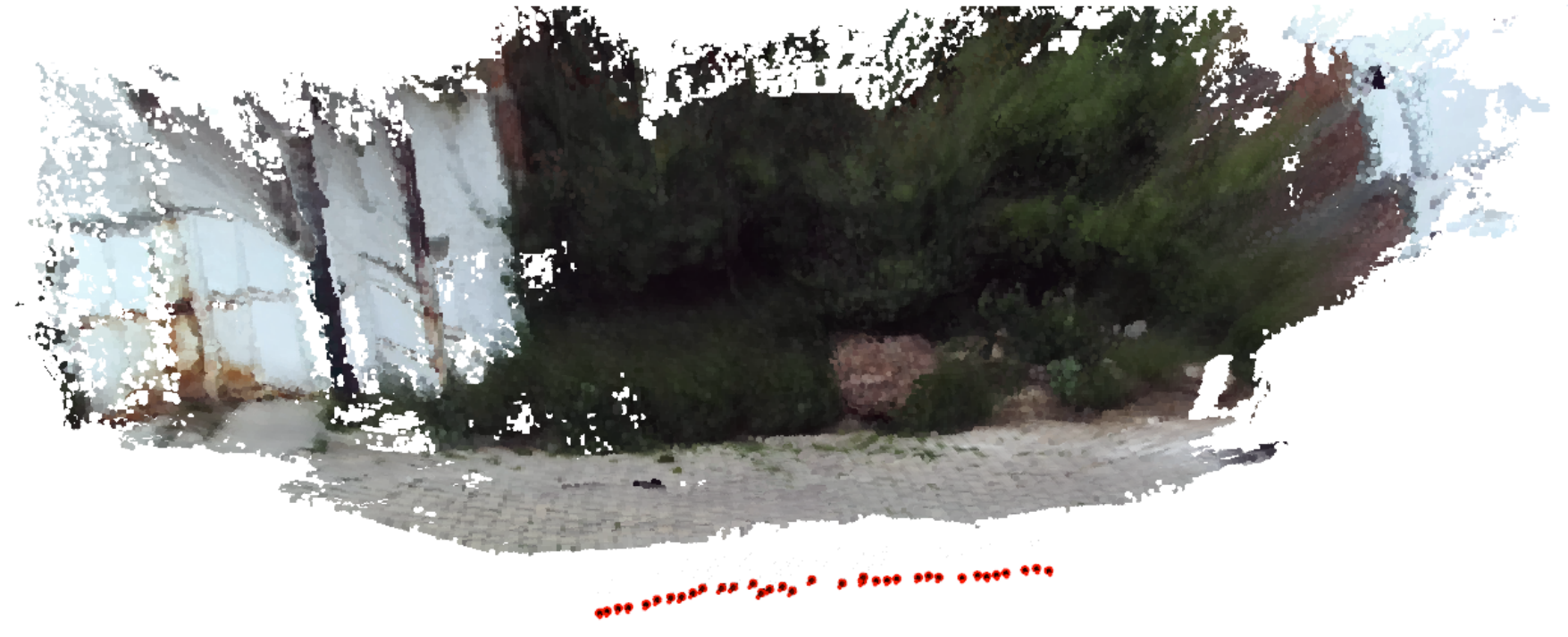}}
    \caption{The STDyn-SLAM when a static object becomes dynamic. Images a) and b) corresponds to the left images from a sequence. Image c) is the 3D reconstruction of the environment; in red dots is the trajectory. The OctoMap node fills empty areas along the sequence of images.}
    \label{fig:ibiza}
\end{figure}

First, we test the STDyn-SLAM in an outdoor environment where a car is parked, and then it moves forward. In this case, a static object (a car) becomes dynamic, see Fig. \ref{fig:ibiza}. This figure shows the 3D reconstruction, where the car appears static in the first images from the sequence, Fig. \ref{fig:ibiza} a). Then, the car became a dynamic object when it moves forward (Fig. \ref{fig:ibiza} b)), so the STDyn-SLAM is capable of filling the empty zone if the scene is covered again, as it is the case in Fig. \ref{fig:ibiza} c).

\begin{figure*}[t]
    \centering
    \begin{subfigure}[b]{\textwidth}
        \includegraphics[width=\textwidth]{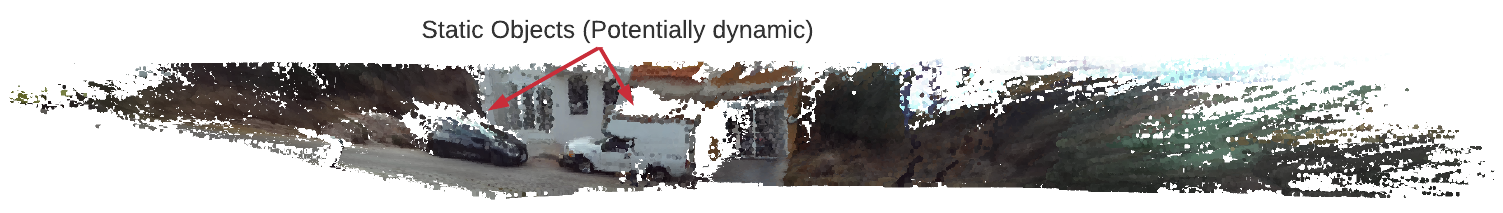}
        \caption{}
        \label{fig:comparison_11}
    \end{subfigure}
    \hfill 
    \begin{subfigure}{\textwidth}
        \includegraphics[width=\textwidth]{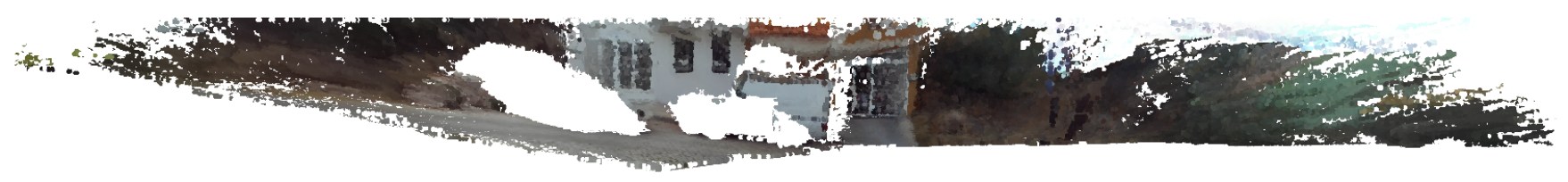}
        \caption{}
        \label{fig:comparison_21}
    \end{subfigure}
    
    \caption{3D reconstruction with the presence of static (two parked cars) and dynamic objects (a person and two dogs). Notice that the person and dogs are not visualized in the scene for the effect of the STDyn-SLAM. Fig. a) depicts the static objects. Nevertheless, the vehicles are potentially dynamic objects, thus in Fig. b), the STDyn-SLAM excludes the bodies considering its possible movement.}
    \label{fig:comparison_empty}
\end{figure*}

The second experiment consists of a scene with two parked cars, a walking person, and a dog. Even though the vehicles are static, the rest of the objects are moving. Fig. \ref{fig:comparison_11} shows the scene taking into account the potentially dynamic entities. However, a car can change its position; the STDyn-SLAM excludes the probable moving bodies (parked cars) to avoid the multiple plotting throughout the reconstruction. This is depicted in Fig. \ref{fig:comparison_21}.

\begin{figure}[t]
    \centering
    \begin{subfigure}[b]{\columnwidth}
        \includegraphics[width=\columnwidth]{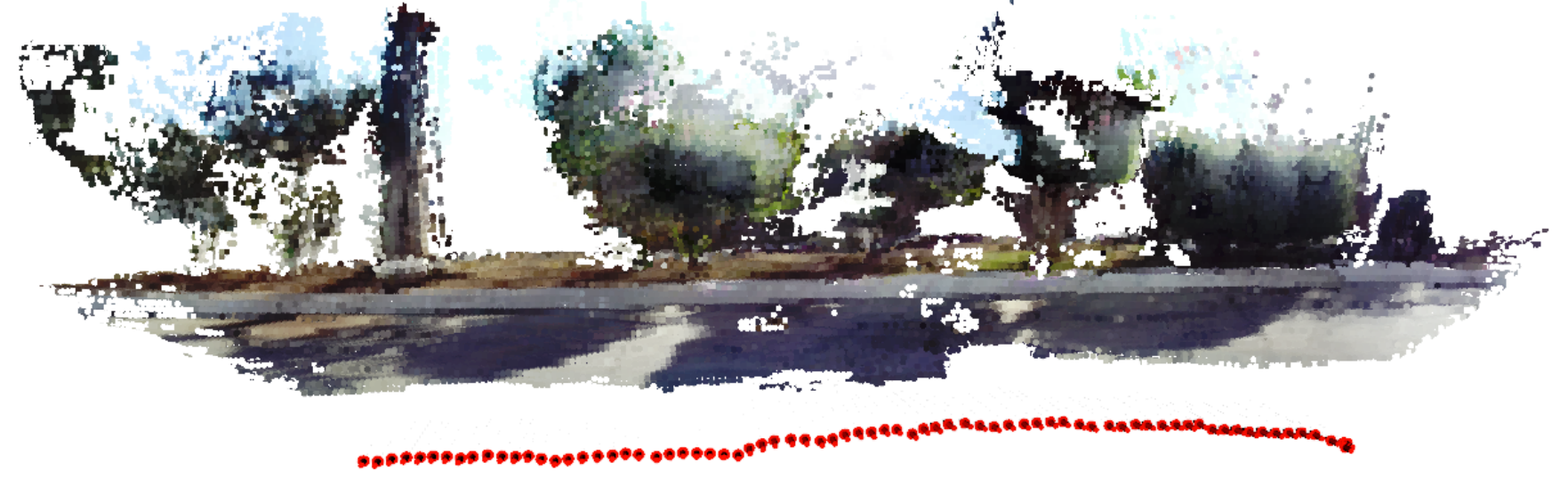}
        \caption{STDyn-SLAM output.}
        \label{fig:comparison_stereo_rtab}
    \end{subfigure}
    \hfill 
    \begin{subfigure}{\columnwidth}
        \includegraphics[width=\columnwidth]{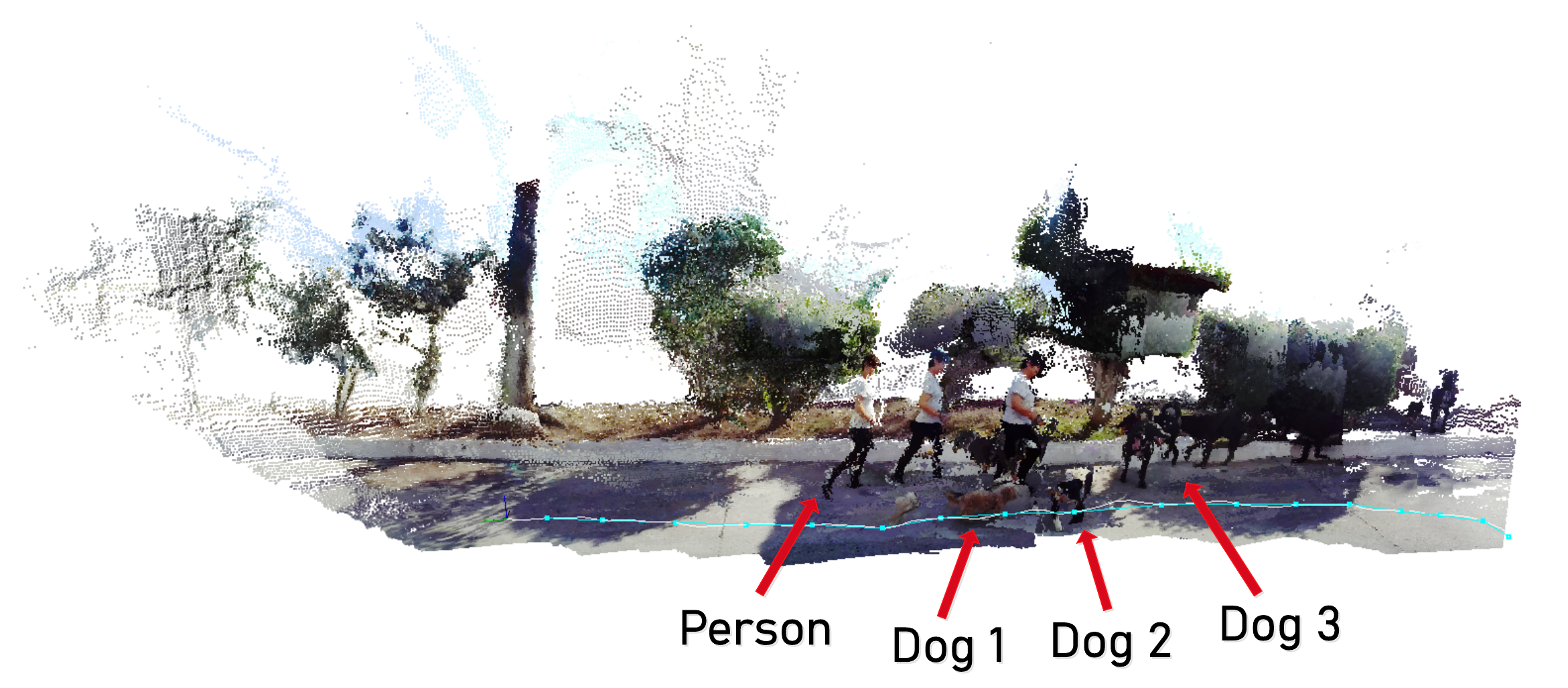}
        \caption{RTAB-Map output.}
        \label{fig:comparison_rtab_blvd}
    \end{subfigure}
    \caption{Experiment comparison between the STDyn-SLAM and the RTABMAP \cite{rtabmap}. Image a) shows the 3D reconstruction given by STDyn-SLAM; it eliminates dynamic objects' effect on the mapping. Image b) shows the point cloud created by RTABMAP; notice how dynamic objects are mapped along the trajectory. This is undesirable behavior.}
    \label{fig:comparison_blvd}
\end{figure}

As a third experiment, we compared the point clouds from the RTABMAP and the STDyn-SLAM systems. The sequence was carried out outdoors with a walking person and two dogs. Since RTABMAP generates a point cloud of the scene, we decided to compare it with our system. To build the 3D reconstructions from RTABMAP, we provided left and depth images, camera info, and odometry as inputs for the RTABMAP. We used stereo and depth images; the intrinsic parameters are saved in a text file in the ORB-SLAM2 package. Fig \ref{fig:comparison_blvd} shows the 3D reconstructions. In Fig \ref{fig:comparison_stereo_rtab} our system excludes the dynamic objects. On the other hand, Fig \ref{fig:comparison_rtab_blvd} RTABMAP plotted the dynamic objects on different sides of the scene resulting in an incorrect map of the environment.
\subsection{Comparison of state-of-art and our SLAM using KITTI datasets}
We compare our SLAM with CubeSLAM and ORBSLAM2 approaches. To evaluate the SLAM systems, we selected sequences with dynamic objects and no-loop closure. Therefore, we chose the 01, 03, 08, and 10 sequences from the odometry KITTI datasets \cite{Geiger2012CVPR}. Furthermore, we employed \textit{EVO} \cite{grupp2017evo} tools to evaluate the Absolute Pose Error, which computes the RMSE (m), mean, $\max$, and $\min$ values.

\begin{figure}[t]
    \centering
    \subcaptionbox{Sequence 01}{\includegraphics[width=0.5\textwidth]{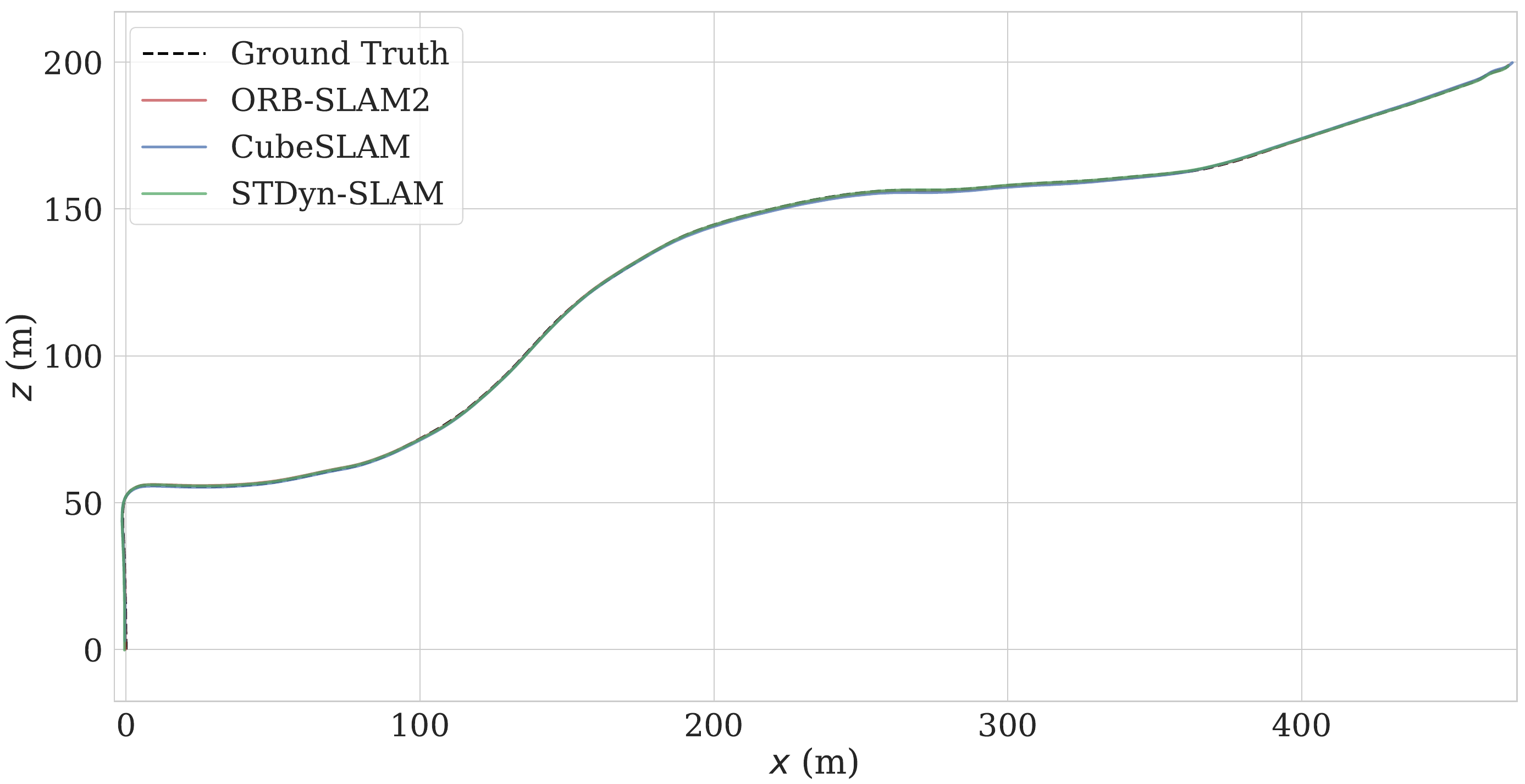}}%
    \\
    \subcaptionbox{Sequence 
    03}{\includegraphics[width=0.5\textwidth]{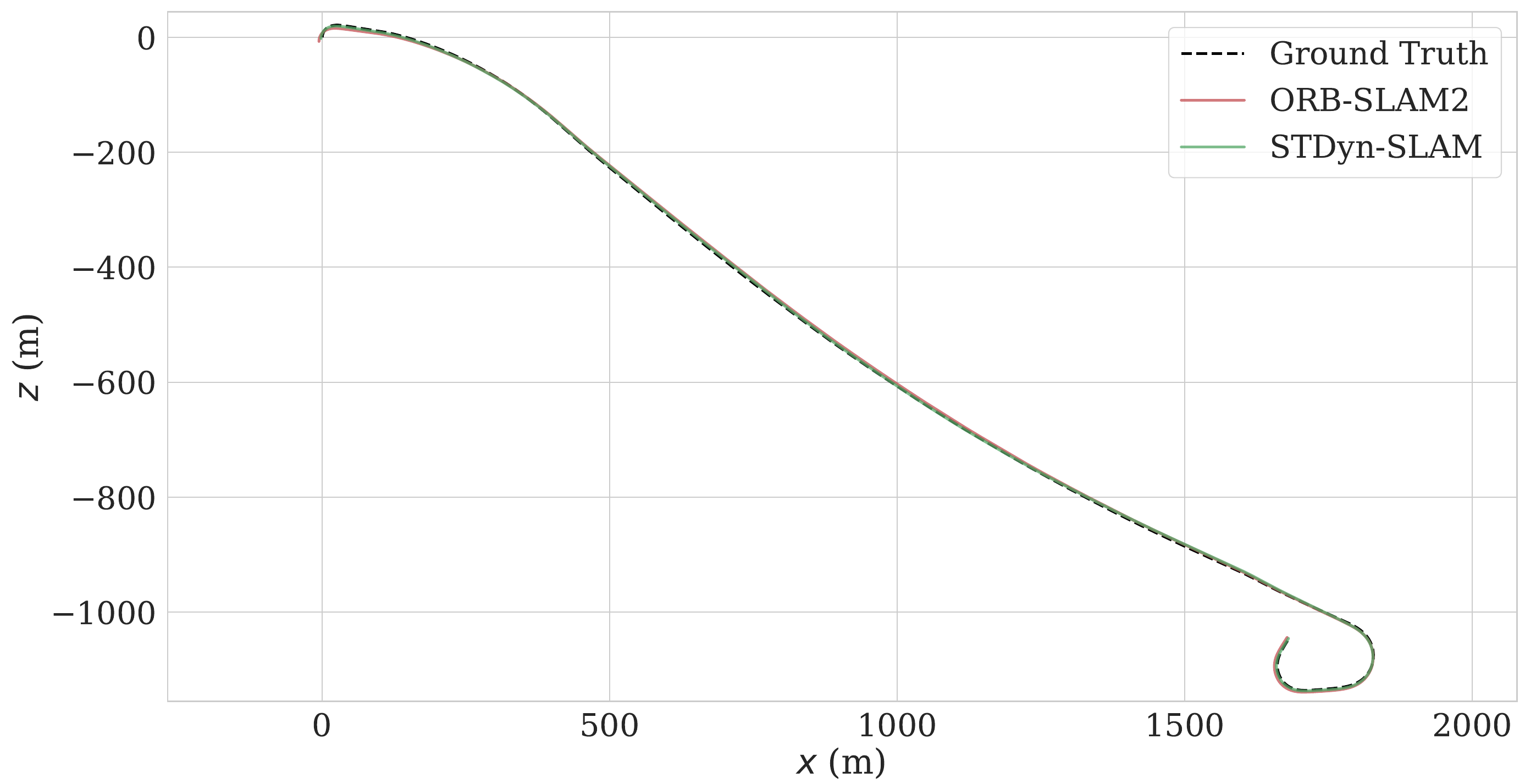}}%
    \\
    \subcaptionbox{Sequence 08}{\includegraphics[width=0.5\textwidth]{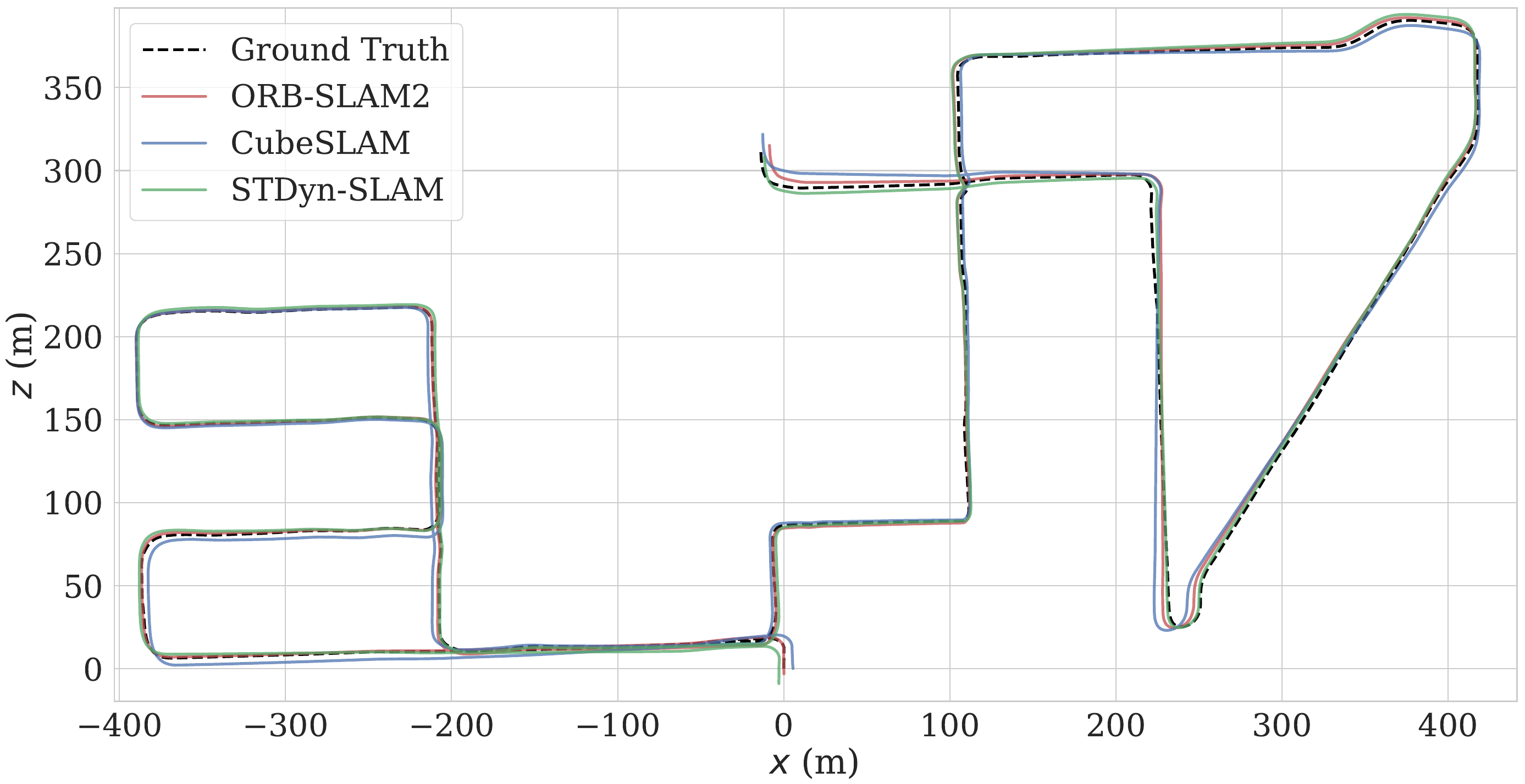}}%
    \\
    \subcaptionbox{Sequence 10}{\includegraphics[width=0.5\textwidth]{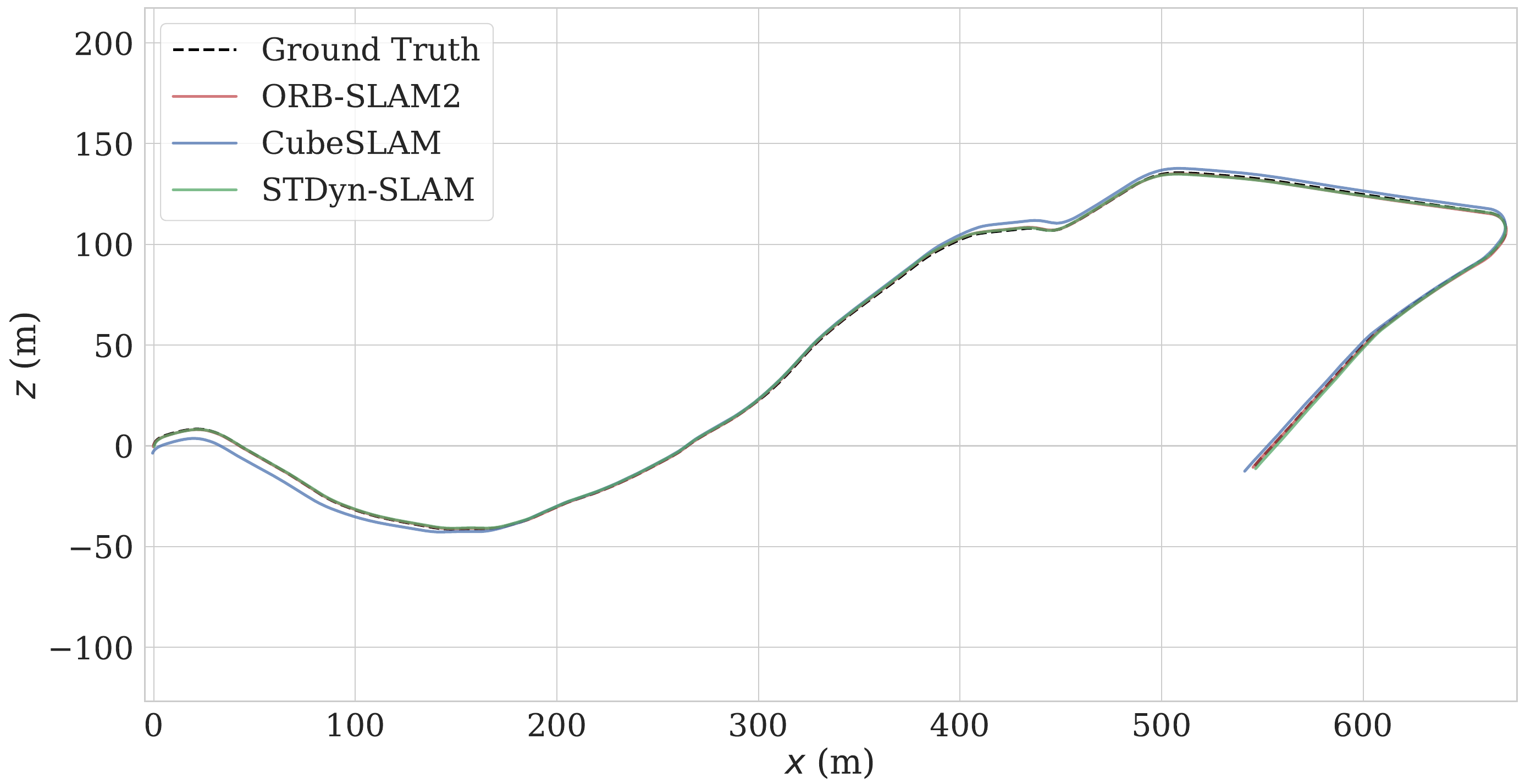}}%
    \caption{Comparison experiments. These figures show the trajectories generated using the KITTI's odometry dataset \cite{Geiger2012CVPR}. The dashed line corresponds to the ground truth, the red line is the ORB-SLAM2 output, the blue line corresponds to the Cube-SLAM, and the green line is the STDyn-SLAM.}
    \label{fig:trajectories}
\end{figure}

Because CubeSLAM is a monocular system, we only took the left images from each sequence to test it. Moreover, we manually completed the trajectory from CubeSLAM since it can not detect all the positions.  Also, we adjusted the path scale of the SLAM systems to coincide with the ground truth. Tab. \ref{tabla:rate} depicts the results of evaluations concerning the ground truth trajectory. We improved the trajectory in the 01 and 08 sequences, where there are dynamic elements. Besides, Fig. \ref{fig:trajectories} shows the comparison between trajectories.

\begin{table*}[t]
    \centering
    \caption{Comparison of Absolute Pose Error (APE) on KITTI Dataset.}
    \begin{tabular}{c || c c c | c c c | c c c | c c c}
        \hline
        Sequence & \multicolumn{3}{c|}{\textbf{rmse}} & \multicolumn{3}{c|}{\textbf{mean}} & \multicolumn{3}{c|}{\textbf{max}} & \multicolumn{3}{c}{\textbf{min}}\\[5pt]
                 &  Cube  &  ORB2  &  Our  &  Cube  &  ORB2  &  Our  &  Cube  &  ORB  & Our  &  Cube  &  ORB2  &  Our \\[5pt]
        \hline
            &         &       &               &         &        &                &        &        &                &
            &         &           \\
        01  &         & 3.810 & \textbf{2.168} &         & 3.469 & \textbf{2.040} &        & 9.068  & \textbf{3.612} &      & 0.918   & \textbf{0.263} \\[5pt]
        
        03  &  0.499  & \textbf{0.256} & 0.268 &  0.468  & \textbf{0.237} & 0.241 & 1.809  & \textbf{0.490}  & 0.841 & 0.120 & 0.062  & \textbf{0.018} \\[5pt]
        
        08  &  4.751  & 3.318 & \textbf{3.224} &  4.018  & 2.796 & \textbf{2.781} & 14.360 & \textbf{9.621} & 13.356 & 0.546 & 0.462 & \textbf{0.112} \\[5pt]

        10  &  1.856  & \textbf{1.020} & 1.227 &  1.600  & \textbf{0.901} & 1.132 & 3.898  & 2.550  & \textbf{2.347} & 0.302 & 0.199  & \textbf{0.091} \\[5pt]
        \hline
    \end{tabular}
    \label{tabla:rate}
\end{table*}
%


\section{Conclusion}\label{conclusion}
This work presents the STDyn-SLAM system for outdoor and even indoor environments where dynamic objects are present. The STDyn-SLAM is based on images captured by a stereo pair for 3D reconstruction of scenes, where the possible dynamic objects are discarded from the map; this allows a trustworthy point cloud. The system capability for computing a reconstruction and localization in real-time depends on the computer's processing power since a GPU is necessary to support the processing. However, with a medium-range computer, the algorithms work correctly

In the future, we plan to implement an optical flow approach based on the last generation of neural networks to improve dynamic object detection. The implementation of neural networks allows replacing classic methods such as geometric constraints. Furthermore, we plan to increase the size of the 3D map to reconstruct larger areas and obtained longer reconstructions of the scenes. The next step is implementing the algorithm in an aerial manipulator constructed in the lab.


\bibliographystyle{IEEEtran}
\bibliography{access.bbl}


\begin{IEEEbiography}[{\includegraphics[width=1in,height=1.25in,clip,keepaspectratio]{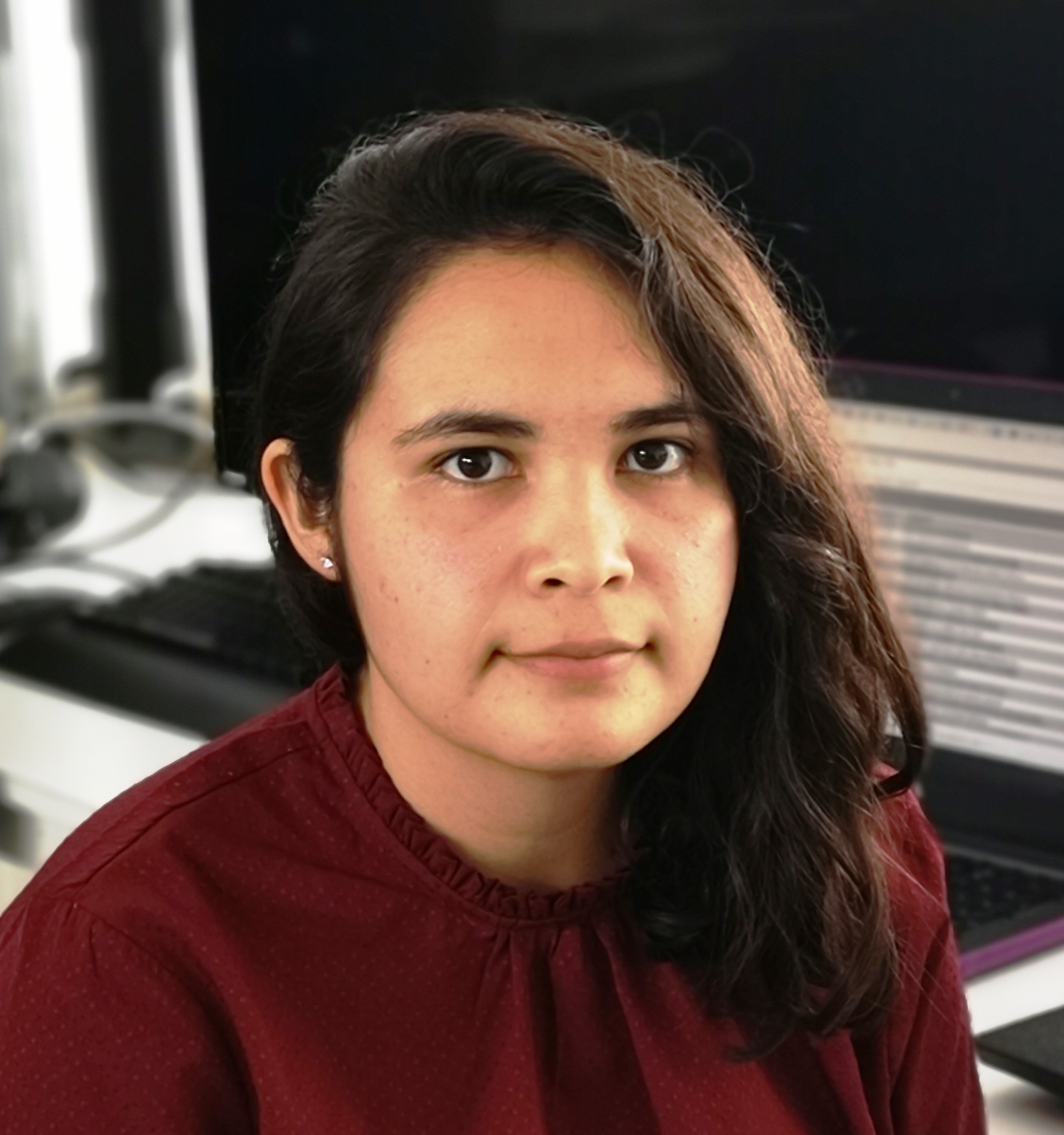}}]{Daniela Esparza} received the B.S. degree in Robotic Engineering from the Universidad Politécnica del Bicentenario, México in 2017. She received the Master's degree in Optomechatronics from the Center for Research in Optics in 2019, where she is currently pursuing a Ph.D. degree in Mechatronics and mechanical design. Her professional interests are focused on artificial vision, such as 3D reconstruction and deep learning applied to SLAM developed over platforms as mobile robots.

\end{IEEEbiography}

\begin{IEEEbiography}[{\vspace{-3ex}\includegraphics[width=1in,height=1.25in,clip,keepaspectratio]{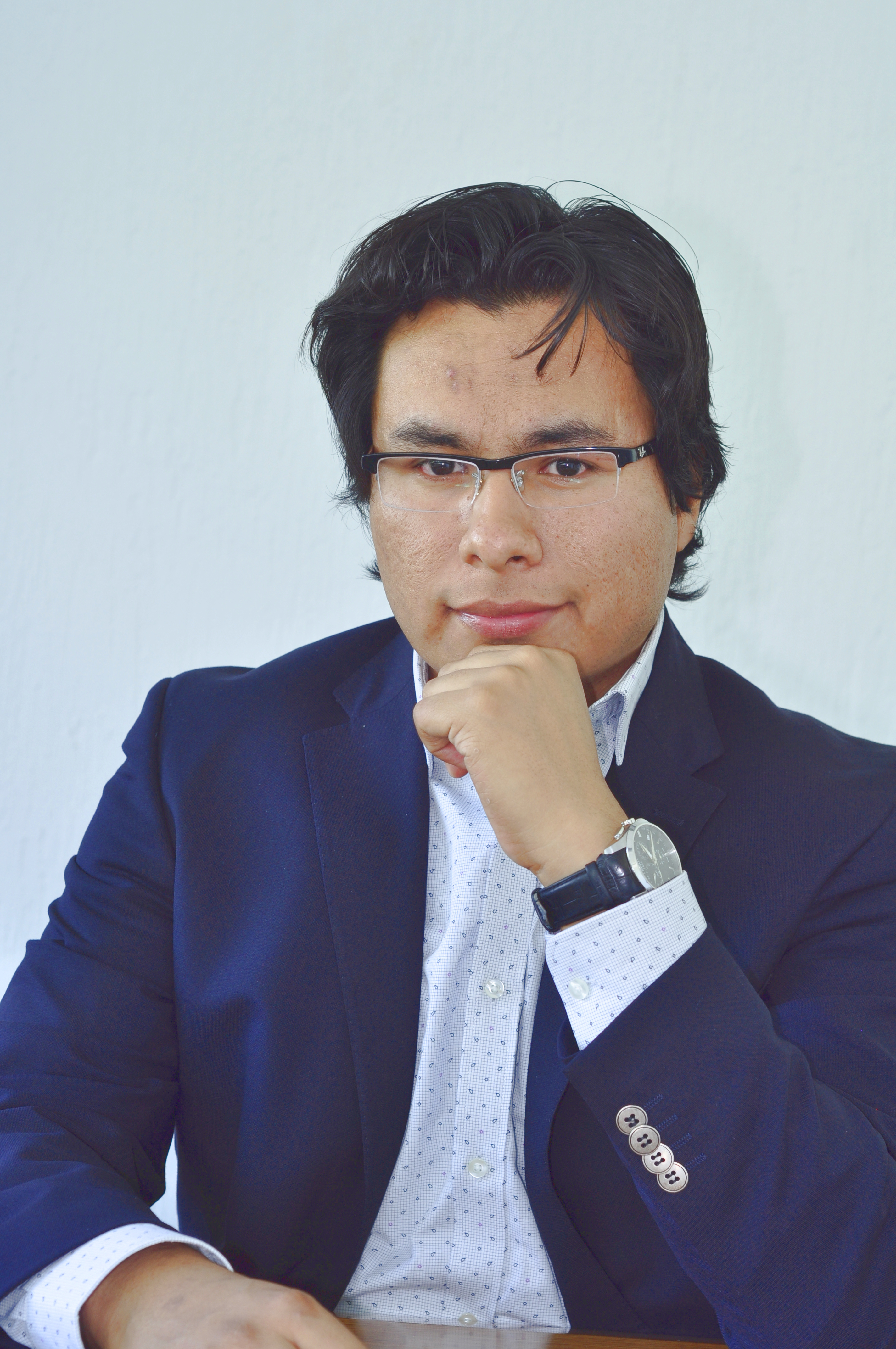}}]
{Gerardo Flores} received the B.S degree in Electronic Engineering with honors from the Instituto Tecnol\'ogico de Saltillo, M\'exico in 2007; the M.S. degree in Automatic Control from CINVESTAV-IPN, Mexico City, in 2010; and the Ph.D. degree in Systems and Information Technology from the Heudiasyc Laboratory of the Universit\'e de Technologie de Compi\`egne - Sorbonne Universit\'es, France in October 2014. Since August 2016, he has been a full-time researcher and Head of the Perception and Robotics LAB with the Center for Research in Optics, Le\'on Guanajuato, Mexico. His current research interests are focused on the theoretical and practical problems arising from the development of autonomous robotic and vision systems. From 2020 he is an Associate Editor of Mathematical Problems in Engineering.
\end{IEEEbiography}
\EOD

\end{document}